# FULL-CHIP CMP MODELING BASED ON FULLY CONVOLUTIONAL NETWORK LEVERAGING WHITE LIGHT INTERFEROMETRY

Jules Exbrayat[1], Renan Bouis[1], Elie Sezestre[1], Viorel Balan[1], Arnaud Cornelis[1], Damien Hebras[1] and Catherine Euvrard[1]
[1] Univ. Grenoble Alpes, CEA, Leti, F-38000 Grenoble, France, renan.bouis@cea.fr

## Abstract

As time-to-market is crucial in the Integrated Circuit (IC) industry, speeding up layout manufacturability verification is essential. Chemical-Mechanical Polishing (CMP) plays a vital role in IC fabrication but is significantly influenced by Layout-Dependent Effects (LDE). An accurate and efficient CMP model enables design teams to correct surface unevenness before fabrication, reducing costs and accelerating the design phase. However, existing models often rely on Density Step Height (DSH) modeling, which is time-consuming for calibration and requires substantial hardware resources for fine-grained predictions.
In this paper, we propose combining the advantages of two surface analysis techniques, White Light Interferometry (WLI) and Atomic Force Microscopy (AFM), to train a deep learning model. This model aims to predict full-chip post-CMP nanotopography with nanometer-scale accuracy. Our deep learning model is based on a Convolutional Neural Network (CNN) and follows a two-step pipeline. The model is trained on each technique separately, resulting in a detailed full-chip CMP model.

## 1 Introduction

CMP is a crucial process in IC semiconductor manufacturing. It supports traditional transistor processes, multilevel interconnections, and advanced techniques such as 3D stacking, 3D packaging, and heterogeneous integration [1].
The primary goal of CMP is to planarize the chip surface, preventing topological issues across layers and avoiding problems like residues and short-circuits. CMP combines chemical and mechanical effects, but its complexity— involving numerous parameters and effects at various scales—makes fine-grained modeling virtually impossible.
Chip design can significantly impact this process, espacially layout density, which may lead to post-CMP topographical effects. Several EDA software providers have developed tools that predict post-CMP surface topography by employing physics-based models [2]. This physical verification of the layout through CMP modeling before IC chip manufacturing is essential for ensuring manufacturability and improving yield.
One of the main challenges of current physics-based CMP modeling is its high computational cost and long inference time. Our solution aims to reduce the time required for CMP modeling by introducing a deep learning model. This model predicts new layout nanotopography by training on layout and surface characterization data obtained after the CMP process.

## 2 CMP Surface Characterization

Layout-based CMP modeling aims to predict topography deviations caused by chip design, such as erosion and dishing. While erosion affects large areas, typically ranging from tens to hundreds of square micrometers, dishing impacts much smaller areas, on the order of several square nanometers. However, both effects result in deviations in the nanometer range in the Z direction. This implies that data measurement accuracy should be at the nanometer scale in the Z direction, while the accuracy in the X and Y directions should match at least the size of the effect. Specifically, the X and Y resolution should be in the micrometer scale for erosion and in the nanometer scale for dishing. To meet these specifications, careful attention must be paid to the measurement techniques used.

### 2.1 White Light Interferometry

To meet the constraints discussed earlier for erosion measurement, White Light Interferometry (WLI) appears to be an excellent choice. WLI enables quick surface topography measurement with a millimeter-scale field of view, micrometer X&Y resolution, and sub-nanometer Z-axis resolution. WLI relies on measuring interference patterns, where a broadband incident light source is split into two paths: one directed to a reference mirror and the other to the sample. This creates an interference pattern known as an interferogram, as illustrated in Figure 1. By precisely monitoring the position of this interferogram along the Z-axis for each measured pixel, it is possible to obtain high-resolution 3D mappings of die surfaces with images containing millions of pixels. This technique has been utilized at LETI for over ten years [3,4].

### 2.2 Atomic Force Microscopy

Similarly, dishing measurement can be effectively performed using Atomic Force Microscopy (AFM), a well-known technique that offers sub-nanometer Z-axis resolution and nanometer-scale X&Y resolution. Its main limitation is the measurement scale, which is limited to dozens of micrometers, making it a local measurement technique. However, this limitation aligns well with the constraints for dishing measurement.

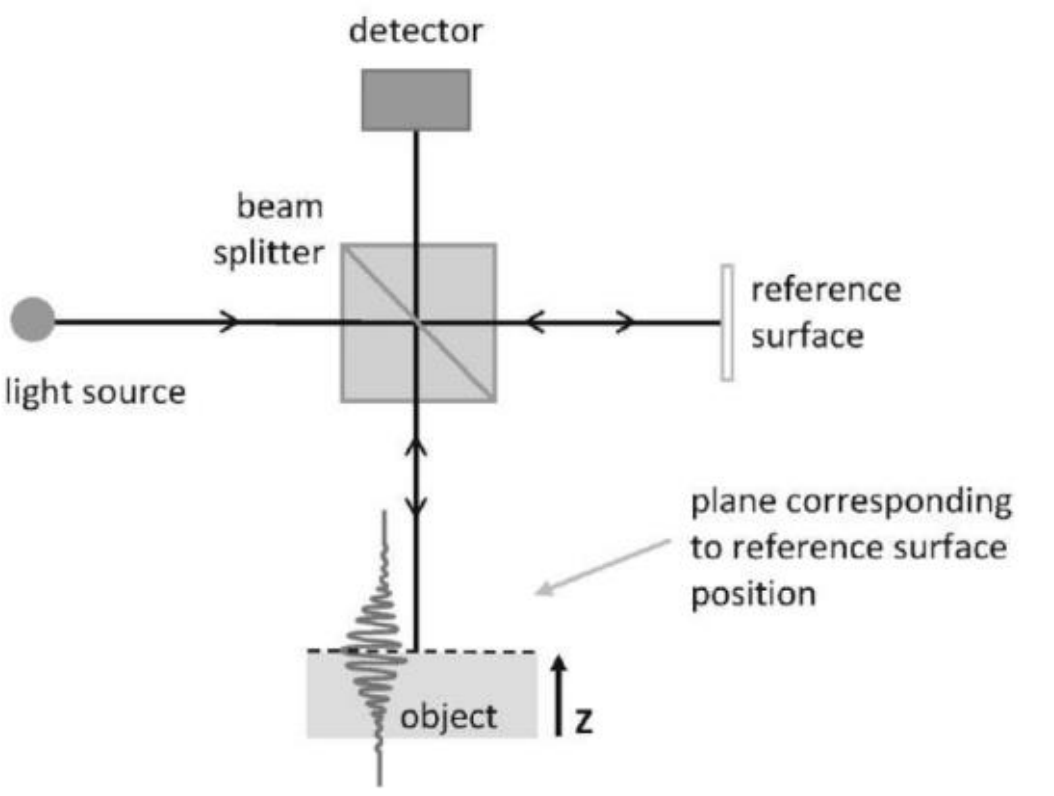

**Figure 1** Schematic view of the WLI technique operating principle (Reproduced from [5])

# 3 CMP Modelling process

Main goal of our work was to experiment the possibility of predicting post CMP process nanotopography of new layout with a nanometer-scale accuracy.

## 3.1 Experimental setup

To achieve this goal, we developed a specific layout (Fig. 2a), designed to represent a typical IC copper level with complex copper patterns in an oxide matrix. Wafers with a single copper level were fabricated at LETI using our layout and processed with our copper CMP process. The full die surface topography of these wafers was characterized using local AFM measurements as well as WLI for full die characterization, resulting in a numerical image of size $h \times w$ (Fig. 2b). To leverage the advantages of both measurement techniques, we constructed a two-stage modeling pipeline. The first stage aims to predict erosion based on WLI data, and the second stage adds dishing prediction based on AFM measurements. The work presented in this paper is focus on predicting erosion as an initial step, while the integration of dishing data will be considered in future work

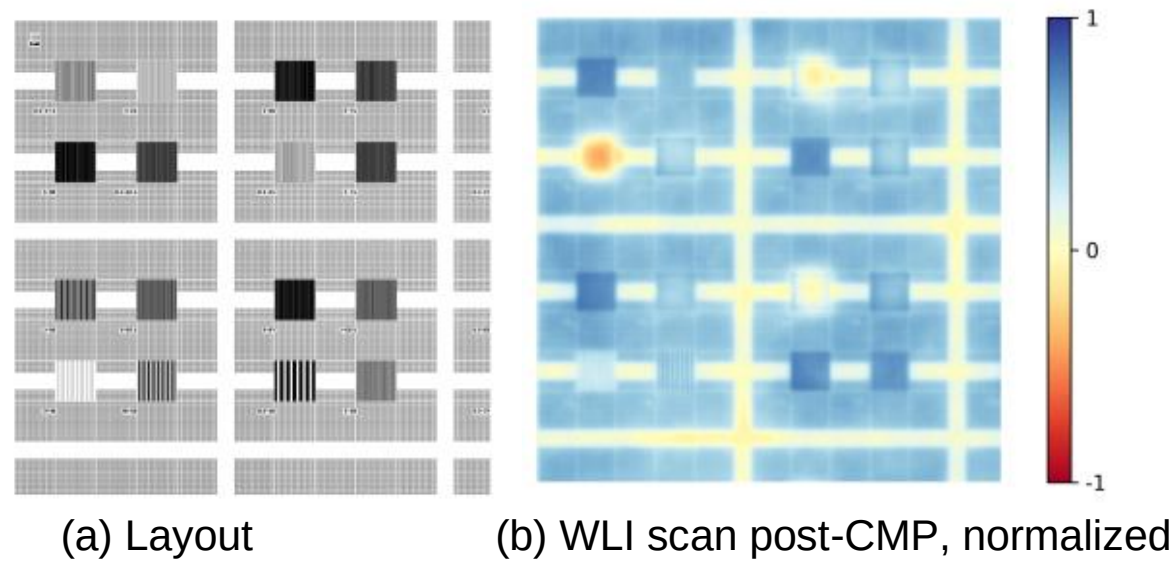

(a) Layout (b) WLI scan post-CMP, normalized

**Figure 2** Input (a) and output (b) of the model, shown as a fraction of the actual data for visualization puposes

## 3.2 Deep learning model

CMP modeling consists in training a neural network with the layout as input data (Fig. 2a) and the corresponding WLI image as output (Fig. 2b).

### 3.2.1 Data preprocessing

#### 3.2.1.1 Layout sampling

To be fed into the network, the vectorial layout file must be sampled at the same resolution as the WLI scan. Sampling data to the correct scale is a core challenge in CMP modeling: while nanometer-scale patterns must be preserved (to allow the model to infer local densities and sharp edges), erosion occurs at the micrometer scale and thus requires a broader context. These differing scales make it difficult to downsample without losing critical details, although downsampling is convenient for reducing computational costs and achieving spatial correspondence between input and output. In our case, we used nearest neighbor interpolation, resulting in a binary layout image (1 for copper, 0 for oxide) of size $h \times w$.

#### 3.2.1.2 Smoothing

As mentioned earlier, we focused on erosion modeling. The WLI scan resolution is not adapted to fully capture dishing variations, resulting in noisy cross-sectional views of the chip topography, as illustrated in Figure 3. Therefore a pre-processing step is required to clean the training set that features only the large-scale erosion signal.

Additionally, since the dishing signal is under-resolved, random spikes may appear in the data, leading to a biased evaluation of the model's loss during both the training and testing phases. To reduce measurement noise caused by dishing artifacts in the WLI scan while preserving the broader erosion patterns, we apply a simple linear smoothing kernel. This kernel averages pixel values over an $m \times n$ neighborhood. This operation effectively suppresses local high-frequency fluctuations and ensures that the model focuses on the relevant physical phenomena.

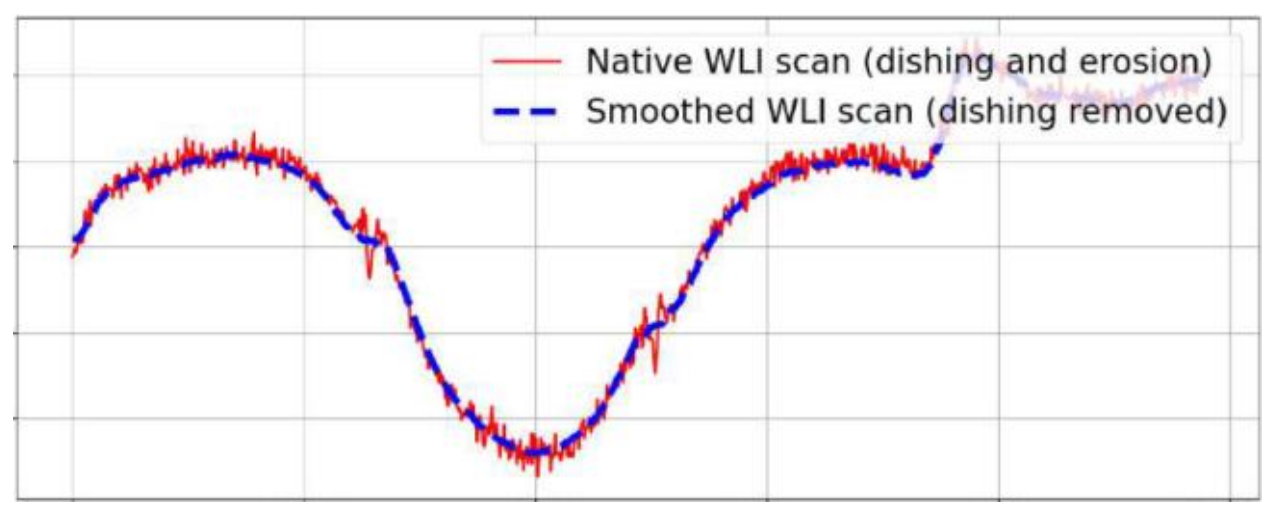

**Figure 3** Cross-sectional view of the smoothed WLI image

3.2.1.3 Normalization

To ensure numerical stability during training and to align the output range with the network's final activation function, min-max normalization was applied to the WLI scans. This normalization is particularly important given the use of the hyperbolic tangent (tanh) activation function at the network's output, as tanh maps values to the range [-1, 1]. This transformation ensures a bounded target range while preserving the shape of the pixel value distribution.

3.2.1.4 Data augmentation

Data augmentation techniques commonly used in computer vision, such as rotations and horizontal and vertical flips (Figure 4), were employed to artificially increase the data volume. This contribute to improve the model's generalization capability. In this work, data augmentation allowed us to increase the dataset size by a factor of eight, thereby enhancing the diversity of training samples while respecting the physical constraints of the problem.

3.2.1.5 Test-train split

To evaluate the model's generalization performance, the dataset was divided into training and testing subsets. Instead of splitting entire dies or scan sequences, individual subframes were randomly selected to compose the test set. This approach ensured a representative sampling of spatial and pattern diversity, preventing overfitting to specific regions. In this work, the test set represented 20% of the total subframes, while the remaining 80% were used for training.

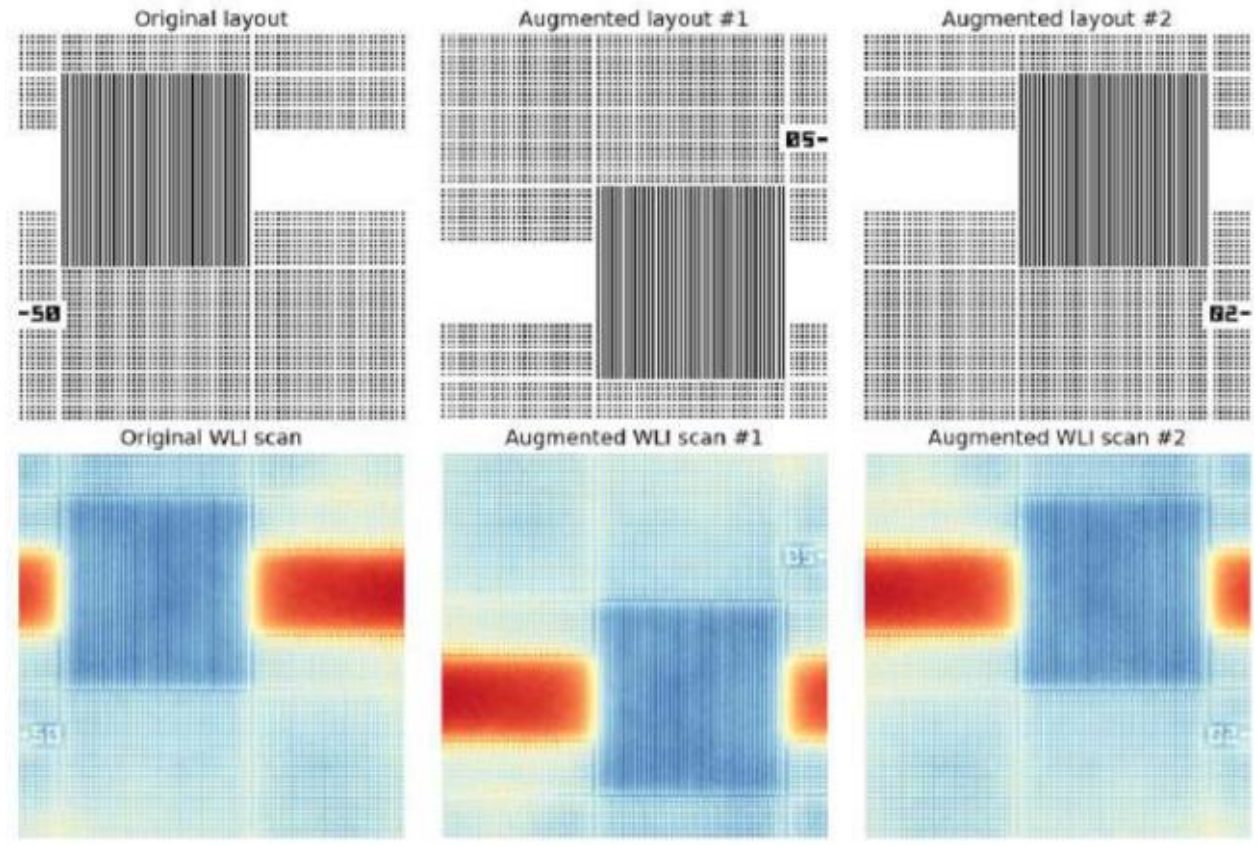

**Figure 4** Example of two data augmentations(center & right) applied to the same data subframe (left).

### 3.2.2 Model architecture

The selected architecture is a Fully Convolutional Network (FCN) based on U-Net, inspired by Yu Ji et al. [6] and represented in Figure 5. It consists of several convolutional layers, with each convolutional kernel learning its own parameters from the data via gradient descent optimization. This contrasts with physics-based models, which rely on explicit equations. Between the convolutional layers, pooling layers perform down-sampling of feature maps, enabling the model to focus on larger scales as it progresses deeper. Specifically, U-Net provides an encoder-decoder structure with skip connections, allowing precise localization by combining high-level features with low-level ones [7].

Similarly, CMP modeling depends on both low-level design features such as copper line corners, edges, and dummies, as well as high-level characteristics like local density, line width, and shapes.

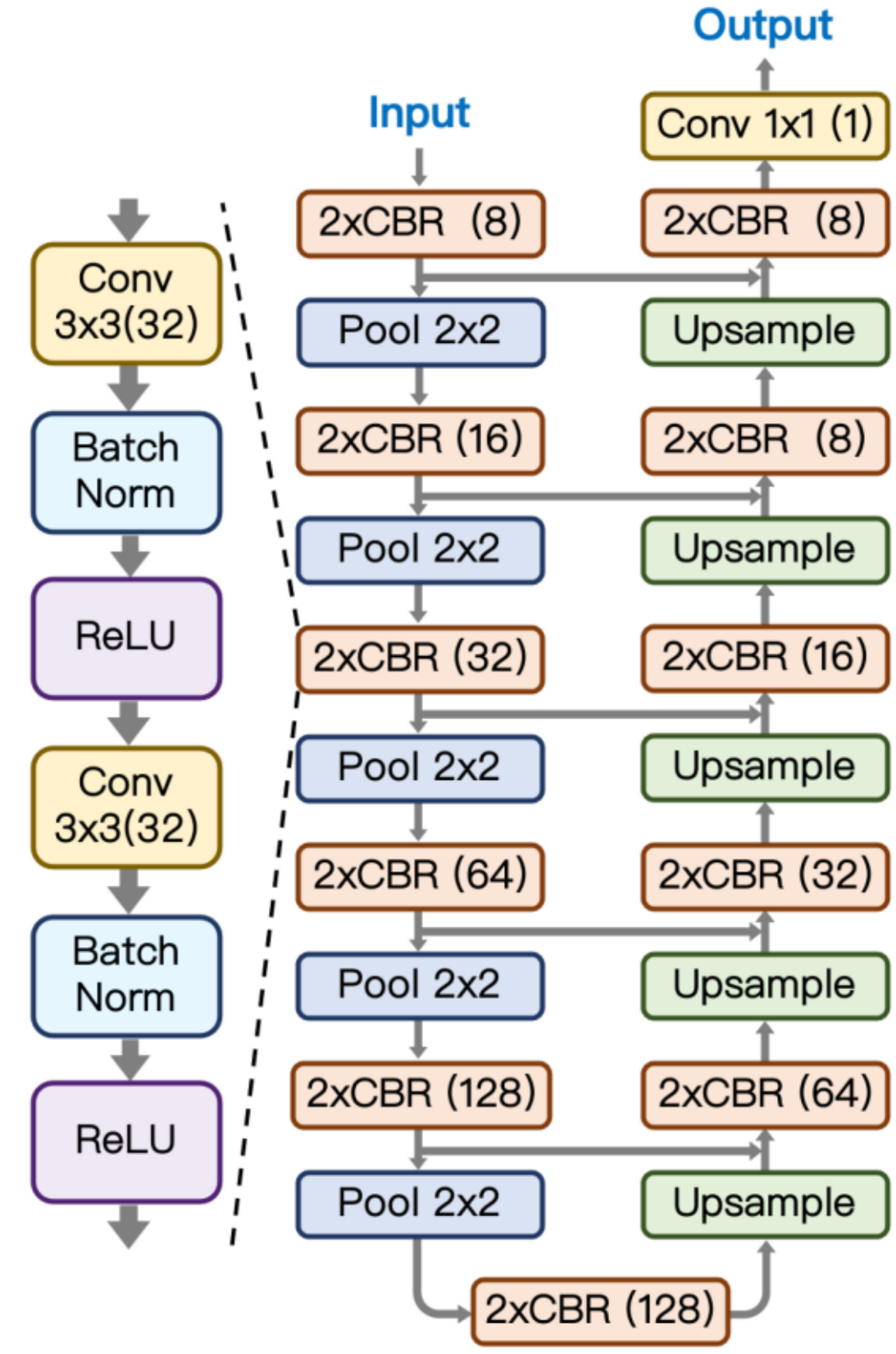

**Figure 5** Architecture of the Fully Convolutional neural Network reproduced from Yu Ji et al.[6]

### 3.2.3 Model convergence

Following data preprocessing treatments, we aimed to minimize the loss between the ground truth and the model's prediction:

$$\min_{\mathbf{w}} \sum_{(x,y)\in\mathcal{D}} \|y - f(x)\|^2$$

where *D* is the dataset, *f* is the model function, *x* is the input, *y* the target output, and *w* denotes the model parameters.

This loss is minimized using gradient descent. At each iteration *t*, the parameters *w* are updated by the following rule:

$$\mathbf{w}_{t+1} = \mathbf{w}_t - \eta \nabla_{\mathbf{w}} \mathcal{L}(\mathbf{w}_t)$$

where $\nabla wL(wt)$ is the loss gradient computed over a minibatch *Bt* and $\eta$ is the learning rate.

## 4 Results & discussions

### 4.1 Model training

The model was trained using PyTorch in Python, with the Adam optimizer and a mean squared error loss function. Training was conducted for up to 150 epochs, with early stopping triggered if there was no improvement in the test loss. The training used a batch size of 16, with 1647 samples of size 128×128 pixels. Evaluation was performed on a separate test set consisting of 412 samples of the same resolution. The best test loss was achieved at the 58th epoch, and the corresponding model weights were saved for final inference.

As shown in Figure 6, the model exhibited rapid initial convergence followed by stable training loss. No signs of overfitting were observed, and in any case, the final model corresponds to the epoch with the lowest test loss. Training was performed on a general-purpose GPU, an Nvidia Tesla V100 with 16GB of memory, with a total training time of approximately 2 hours for the full 150 epochs.

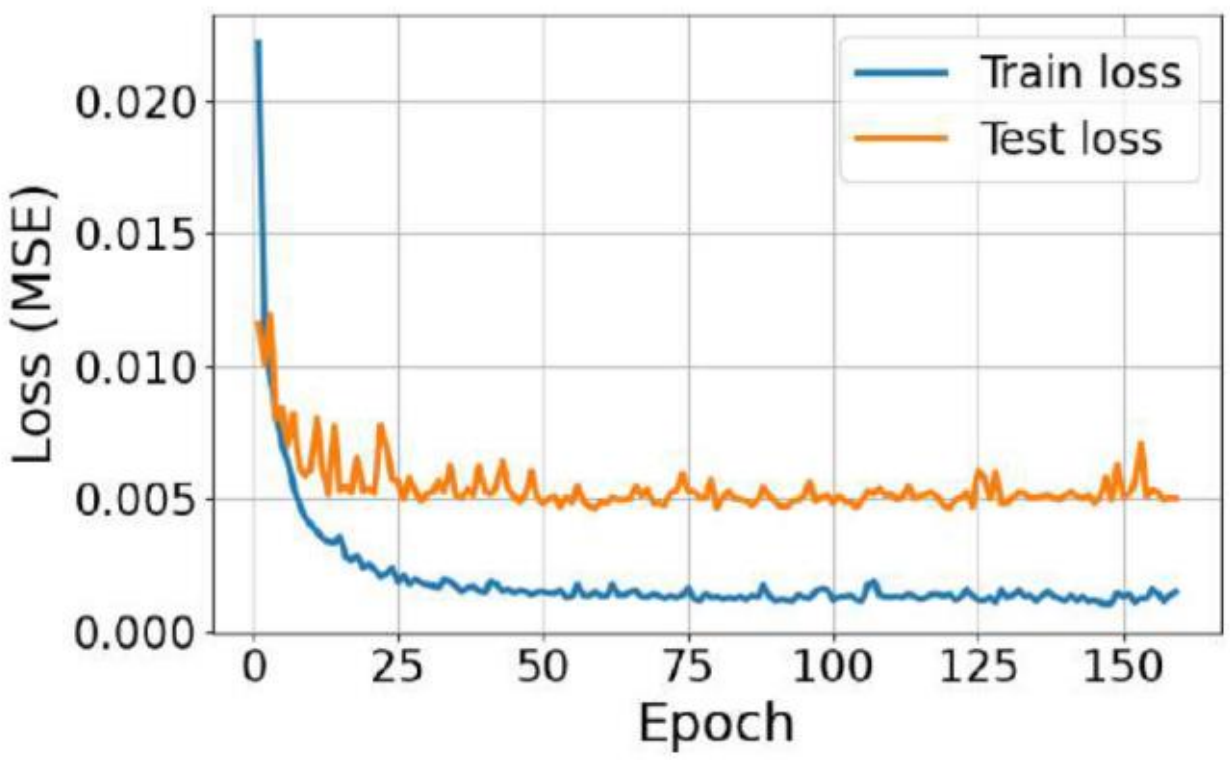

**Figure 6** Training and test loss curves over epochs

### 4.2 Model evaluation

We evaluated various models quantitatively and qualitatively. Qualitatively, the ground truth versus the prediction can be visualized as a color-scaled image for the full chip (not shown) or as a cross-sectional topography curve (Fig. 7). In either case, careful attention must be paid to CMP hotspots, such as pattern borders, corners, and dense copper regions.

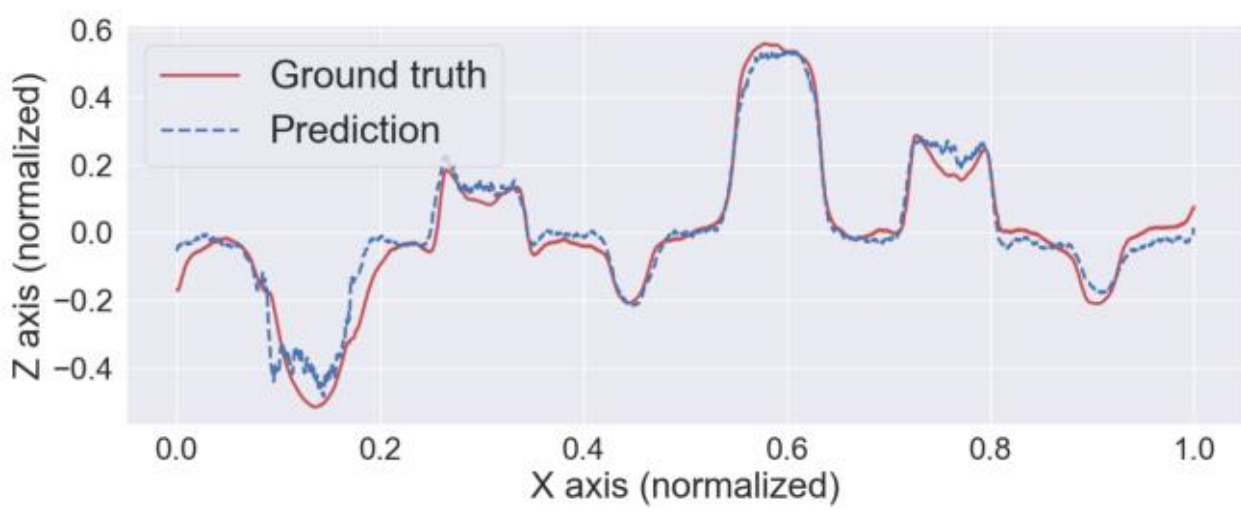

**Figure 7** Cross-sectional view of chip topography, comparing the model's prediction (blue dotted line) to the measured ground truth (red line).

Quantitatively, the model's accurary was evaluated using the L1 loss (mean absolute error) and the root mean squared error (RMSE) where $h \cdot w$ is the image size and $D$ is the test dataset size:

$$L_1 = \frac{1}{|\mathcal{D}|} \sum_{(x,y)\in\mathcal{D}} \frac{\|y - f(x)\|}{h \cdot w}$$

$$\text{RMSE} = \sqrt{\frac{1}{|\mathcal{D}|} \sum_{(x,y)\in\mathcal{D}} \frac{\|y - f(x)\|^2}{h \cdot w}}$$

Table 1 shows that our optimized model achieved strong accuracy compared to state-of-the-art fine-grained CMP modeling by Yu Ji et al. [6] for both L1 and RMSE values. Our model also delivered a fast inference time of 8.1 seconds per sample, significantly outperforming physics-based models, which typically require several hours for a full chip. However, this speed comes at the cost of an initial training effort (see Section 4.1), which takes a few hours on a GPU. In contrast, traditional models do not require training but are computationally intensive at inference time.

| Metric | L1 (nm) | RMSE (nm) | Inference time (s) |
|---|---|---|---|
| Yu Ji et al. (2023) | 1.06 | 1.49 | 25.0 |
| Ours | **0.52** | **0.73** | **8.1** |

**Table 1** Model performance on test data, unseen during training.

## 5 Conclusion

Within the context of CEA's research, this work is viewed as a first step toward predicting the post-CMP process nanotopography of new layouts with nanometer-scale accuracy. We have demonstrated the feasibility of using deep learning techniques coupled with advanced characterization techniques like WLI to achieve fast and reliable full-chip CMP modeling and erosion prediction. Future work will focus on incorporating dishing values, measured by AFM, into the two-step modeling pipeline to obtain accurate full-chip nanometer-scale topography predictions.